\documentclass[11pt,a4paper]{article}
\usepackage[hyperref]{acl2020}
\usepackage{times}
\usepackage{latexsym}

\usepackage{url}
\usepackage{amsmath}
\usepackage{amssymb}
\usepackage{amsfonts}
\usepackage{microtype}
\usepackage{xfrac}
\usepackage{graphicx}
\usepackage{booktabs}
\usepackage{adjustbox}
\usepackage{multirow}
\usepackage{array}
\usepackage{subfig}

\usepackage{cleveref}
\crefname{section}{\S}{\S\S}
\Crefname{section}{\S}{\S\S}
\crefname{table}{Tab.}{}
\crefname{figure}{Figure}{}
\crefname{algorithm}{Algorithm}{}
\crefname{equation}{equation}{}
\crefname{appendix}{Appendix}{}
\crefformat{section}{\S#2#1#3}  %

\newcommand{\insertAllTable}{
\begin{table*}[h!]
\begin{center}
\resizebox{1\linewidth}{!}{
\begin{tabular}[b]{c|ccccc|cccc|cccc|cccc}
\toprule
\multirow{2}{*}{Model} & \multirow{2}{*}{Domain} & \multirow{2}{*}{BPE Merges} & \multirow{2}{*}{Anchors Pts} & \multirow{2}{*}{Share Param.} & \multirow{2}{*}{Softmax} & \multicolumn{4}{c|}{XNLI (Acc)} &
\multicolumn{4}{c|}{NER (F1)} &
\multicolumn{4}{c}{Parsing (LAS)} \\
& & & & & & fr & ru & zh & $\Delta$ & fr & ru & zh & $\Delta$ & fr & ru & zh & $\Delta$ \\
\midrule
\textbf{Default} & Wiki-Wiki & 80k & all & all & shared & 73.6 & 68.7 & 68.3 & 0.0 & 79.8 & 60.9 & 63.6 & 0.0 & 73.2 & 56.6 & 28.8 & 0.0 \\
\midrule 
\multicolumn{15}{l}{\textit{Domain Similarity} (\cref{sec:domain-similarity})} \\
\midrule
\textbf{Wiki-CC} & Wiki-CC & - & - & - & - & 74.2 & 65.8 & 66.5 & -1.4 & 74.0 & 49.6 & 61.9 & -6.2 & 71.3 & 54.8 & 25.2 & -2.5 \\
\midrule 
\multicolumn{15}{l}{\textit{Anchor Points} (\cref{sec:anchor-points})} \\
\midrule
\textbf{No anchors} & - & 40k/40k & 0 & - & - & 72.1 & 67.5 & 67.7 & -1.1 & 74.0 & 57.9 & 65.0 & -2.4 & 72.3 & 56.2 & 27.4 & -0.9 \\
\textbf{Default anchors} & - & 40k/40k & - & - & - & 74.0 & 68.1 & 68.9 & +0.1 & 76.8 & 56.3 & 61.2 & -3.3 & 73.0 & 57.0 & 28.3 & -0.1 \\
\textbf{Extra anchors} & - & - & extra & - & - & 74.0 & 69.8 & 72.1 & +1.8 & 76.1 & 59.7 & 66.8 & -0.5 & 73.3 & 56.9 & 29.2 & +0.3 \\
\midrule 
\multicolumn{15}{l}{\textit{Parameter Sharing} (\cref{sec:param-sharing})} \\
\midrule
\textbf{Sep Emb} & - & 40k/40k & 0* & Sep Emb & lang-specific & 72.7 & 63.6 & 60.8 & -4.5 & 75.5 & 57.5 & 59.0 & -4.1 & 71.7 & 54.0 & 27.5 & -1.8 \\
\textbf{Sep L1-3} & - & 40k/40k & - & Sep L1-3 & - & 72.4 & 65.0 & 63.1 & -3.4 & 74.0 & 53.3 & 60.8 & -5.3 & 69.7 & 54.1 & 26.4 & -2.8 \\
\textbf{Sep L1-6} & - & 40k/40k & - & Sep L1-6 & - & 61.9 & 43.6 & 37.4 & -22.6 & 61.2 & 23.7 & 3.1 & -38.7 & 61.7 & 31.6 & 12.0 & -17.8 \\
\textbf{Sep Emb + L1-3} & - & 40k/40k & 0* & Sep Emb + L1-3 & lang-specific & 69.2 & 61.7 & 56.4 & -7.8 & 73.8 & 46.8 & 53.5 & -10.0 & 68.2 & 53.6 & 23.9 & -4.3 \\
\textbf{Sep Emb + L1-6} & - & 40k/40k & 0* & Sep Emb + L1-6 & lang-specific & 51.6 & 35.8 & 34.4 & -29.6 & 56.5 & 5.4 & 1.0 & -47.1 & 50.9 & 6.4 & 1.5 & -33.3 \\
\bottomrule
\end{tabular}
}
\caption{Dissecting bilingual MLM based on zero-shot cross-lingual transfer performance. - denote the same as the first row (\textbf{Default}). $\Delta$ denote the difference of average task performance between a model and \textbf{Default}.
\label{tab:all}}
\end{center}
\vspace{-0.4cm}
\end{table*}
}

\aclfinalcopy %

\title{Emerging Cross-lingual Structure in Pretrained Language Models}

\author{
  Shijie Wu$^\spadesuit$\thanks{\ \ Equal contribution. Work done while Shijie was interning at Facebook AI.} \space\space\space
  Alexis Conneau$^\heartsuit$\footnotemark[1] \space\space\space \\
\bf Haoran Li$^\heartsuit$ \space\space\space
  Luke Zettlemoyer$^\heartsuit$ \space\space\space
  Veselin Stoyanov$^\heartsuit$ \\
  $^\spadesuit$Department of Computer Science, Johns Hopkins University \\
  $^\heartsuit$Facebook AI\\
  {\tt shijie.wu@jhu.edu}, \space\space\space
  {\tt aconneau@fb.com}\\
  {\tt \{aimeeli,lsz,ves\}@fb.com}
  }

\date{}

\begin{document}
\maketitle
\begin{abstract}
We study the problem of multilingual masked language modeling, i.e. the training of a single model on concatenated text from multiple languages, and present a detailed study of several factors that influence why these models are so effective for cross-lingual transfer. We show, contrary to what was previously hypothesized, that transfer is possible even when there is no shared vocabulary across the monolingual corpora and also when the text comes from very different domains. The only requirement is that there are some shared parameters in the top layers of the multi-lingual encoder. To better understand this result, we also show that representations from monolingual BERT models in different languages can be aligned post-hoc quite effectively, strongly suggesting that, much like for non-contextual word embeddings, there are universal latent symmetries in the learned embedding spaces. For multilingual masked language modeling, these symmetries are automatically discovered and aligned during the joint training process. 
\end{abstract}

\section{Introduction}
Multilingual language models such as mBERT \cite{devlin-etal-2019-bert} and XLM \cite{lample2019cross} enable effective cross-lingual transfer --- it is possible to learn a model from supervised data in one language and apply it to another with no additional training.
Recent work has shown that transfer is effective for a wide range of tasks~\cite{wu-dredze-2019-beto,pires-etal-2019-multilingual}. These work speculates why multilingual pretraining works (e.g. shared vocabulary), but only experiment with a single reference mBERT and is unable to systematically measure these effects.

In this paper, we present the first detailed empirical study of the effects of different masked language modeling (MLM) pretraining regimes on cross-lingual transfer. Our first set of experiments is a detailed ablation study on a range of zero-shot cross-lingual transfer tasks.
Much to our surprise, we discover that language universal representations emerge in pretrained models without the requirement of any shared vocabulary or domain similarity, and even when only a subset of the parameters in the joint encoder are shared.
In particular, by systematically varying the amount of shared vocabulary between two languages during pretraining, we show that the amount of overlap only accounts for a few points of performance in transfer tasks, much less than might be expected. 
By sharing parameters alone, pretraining learns to map similar words and sentences to similar hidden representations. 

To better understand these effects, we also analyze multiple monolingual BERT models trained independently. We find that monolingual models trained in different languages learn representations that align with each other surprisingly well, even though they have no shared parameters. This result closely mirrors the widely observed fact that word embeddings can be effectively aligned across languages~\cite{mikolov2013exploiting}. Similar dynamics are at play in MLM pretraining, and at least in part explain why they aligned so well with relatively little parameter tying in our earlier experiments. 

This type of emergent language universality has interesting theoretical and practical implications. We gain insight into why the models transfer so well and open up new lines of inquiry into what properties emerge in common in these representations. They also suggest it should be possible to adapt pretrained models to new languages with little additional training and it may be possible to better align independently trained representations without having to jointly train on all of the (very large) unlabeled data that could be gathered. For example, concurrent work has shown that a pretrained MLM model can be rapidly fine-tuned to another language \cite{artetxe2019cross}.

This paper offers the following contributions:
\begin{itemize}
\item We provide a detailed ablation study on cross-lingual representation of bilingual BERT.
We show parameter sharing plays the most important role in learning cross-lingual representation, while shared BPE, shared softmax and domain similarity play a minor role.
\item We demonstrate even without any shared subwords (anchor points) across languages, cross-lingual representation can still be learned. With bilingual dictionary, we propose a simple technique to create more anchor points by creating synthetic code-switched corpus, benefiting especially distantly-related languages.
\item We show monolingual BERTs of different language are similar with each other. Similar to word embeddings \cite{mikolov2013exploiting}, we show monolingual BERT can be easily aligned with linear mapping to produce cross-lingual representation space at each level.
\end{itemize}

\section{Background}

\paragraph{Language Model Pretraining}
Our work follows in the recent line of language model pretraining.
ELMo \cite{peters-etal-2018-deep} first popularized representation learning from a language model. The representations are used in a transfer learning setup to improve performance on a variety of downstream NLP tasks. Follow-up work by \newcite{howard-ruder-2018-universal,radford2018improving} further improves on this idea by fine-tuning the entire language model. BERT \cite{devlin-etal-2019-bert} significantly outperforms these methods by introducing a masked-language model and next-sentence prediction objectives combined with a bi-directional transformer model.

The multilingual version of BERT (dubbed mBERT) trained on Wikipedia data of over 100 languages obtains strong performance on zero-shot cross-lingual transfer without using any parallel data during training \cite{wu-dredze-2019-beto,pires-etal-2019-multilingual}. This shows that multilingual representations can emerge from a shared Transformer with a shared subword vocabulary. Cross-lingual language model (XLM) pretraining \cite{lample2019cross} was introduced concurrently to mBERT. On top of multilingual masked language models, they investigate an objective based on parallel sentences as an explicit cross-lingual signal. XLM shows that cross-lingual language model pretraining leads to a new state of the art on XNLI~\cite{conneau-etal-2018-xnli}, supervised and unsupervised machine translation~\cite{lample-etal-2018-phrase}.
Other work has shown that mBERT outperforms word embeddings on token-level NLP tasks ~\cite{wu-dredze-2019-beto}, and that adding character-level information \cite{mulcaire-etal-2019-polyglot} and using multi-task learning~\cite{huang-etal-2019-unicoder} can improve cross-lingual performance.

\paragraph{Alignment of Word Embeddings}
Researchers working on word embeddings noticed early that embedding spaces tend to be shaped similarly across different languages \cite{mikolov2013exploiting}. This inspired work in aligning monolingual embeddings. The alignment was done by using a bilingual dictionary to project words that have the same meaning close to each other \cite{mikolov2013exploiting}. This projection aligns the words outside of the dictionary as well due to the similar shapes of the word embedding spaces. Follow-up efforts only required a very small seed dictionary (e.g., only numbers \cite{artetxe-etal-2017-learning}) or even no dictionary at all \cite{conneau2017word,zhang-etal-2017-adversarial}. Other work has pointed out that word embeddings may not be as isomorphic as thought \cite{sogaard-etal-2018-limitations}  especially for distantly related language pairs \cite{patra-etal-2019-bilingual}. \newcite{ormazabal-etal-2019-analyzing} show joint training can lead to more isomorphic word embeddings space.

\newcite{schuster-etal-2019-cross} showed that ELMo embeddings can be aligned by a linear projection as well. They demonstrate a strong zero-shot cross-lingual transfer performance on dependency parsing. \newcite{wang-etal-2019-cross} align mBERT representations and evaluate on dependency parsing as well.

\paragraph{Neural Network Activation Similarity}
We hypothesize that similar to word embedding spaces, language-universal structures emerge in pretrained language models. While computing word embedding similarity is relatively straightforward, the same cannot be said for the deep contextualized BERT models that we study.
Recent work introduces ways to measure the similarity of neural network activation between different layers and different
models \cite{laakso2000content,li2016convergent,raghu2017svcca,morcos2018insights,wang2018towards}. For example, \newcite{raghu2017svcca} use canonical correlation analysis (CCA) and a new method, singular vector canonical correlation analysis (SVCCA), to show that early layers converge faster than upper layers in convolutional neural networks. \newcite{kudugunta-etal-2019-investigating} use SVCCA to investigate the multilingual representations obtained by the encoder of a massively multilingual neural machine translation system \cite{aharoni-etal-2019-massively}. \newcite{kornblith2019similarity} argues that CCA fails to measure meaningful similarities between representations that have a higher dimension than the number of data points and introduce the centered kernel alignment (CKA) to solve this problem. They successfully use CKA to identify correspondences between activations in networks trained from different initializations.

\begin{figure*}[t]
	\begin{minipage}[t]{0.47\linewidth}
	\begin{center}
        \includegraphics[scale=0.11]{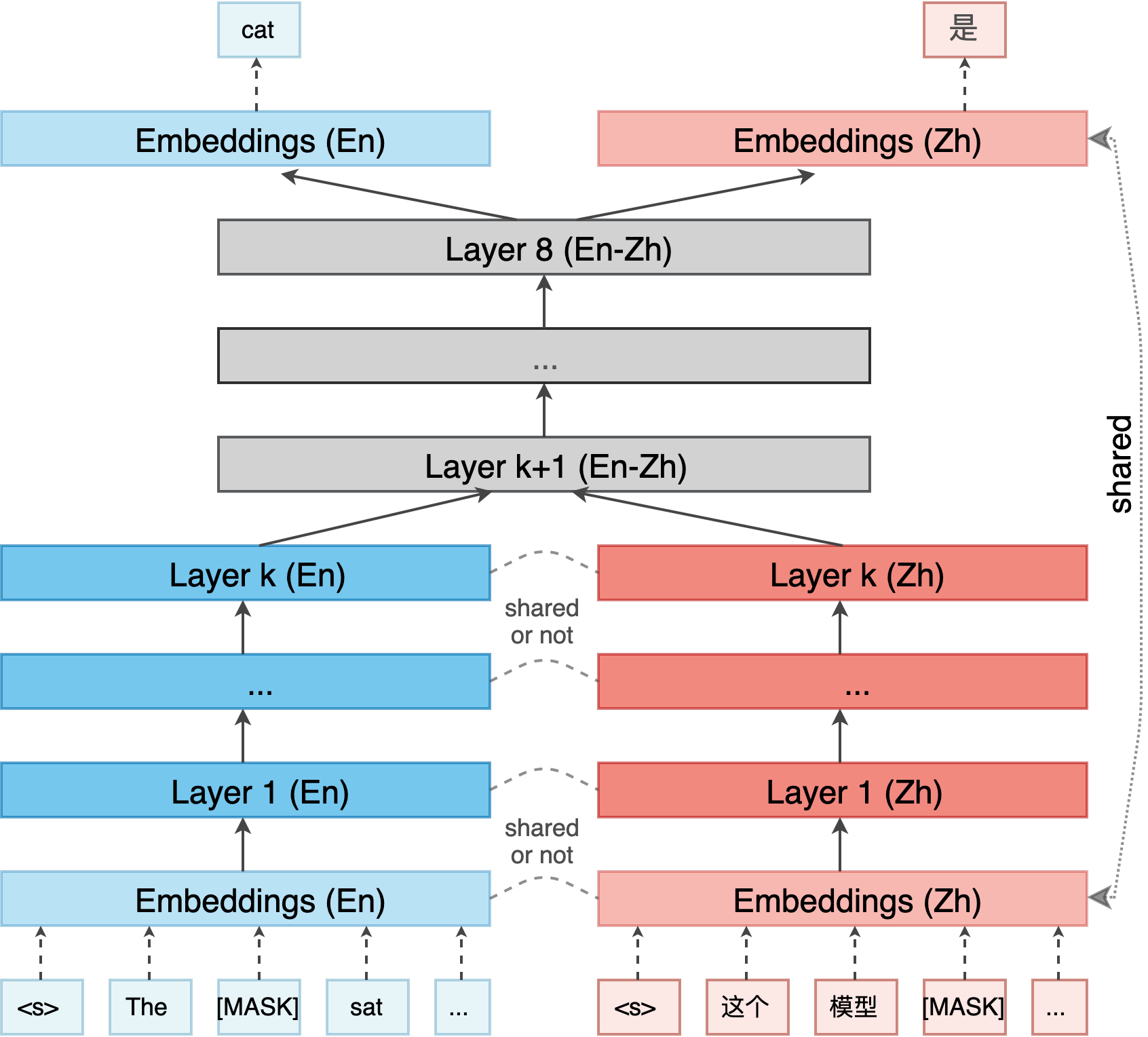}
        \caption{\small \textbf{On the impact of anchor points and parameter sharing on the emergence of multilingual representations.} We train bilingual masked language models and remove parameter sharing for the embedding layers and first few Transformers layers to probe the impact of anchor points and shared structure on cross-lingual transfer.}
            \label{fig:params-1}
	\end{center}
    \end{minipage}
    \hfill
	\begin{minipage}[t]{0.47\linewidth}
	\begin{center}
        \includegraphics[scale=0.13]{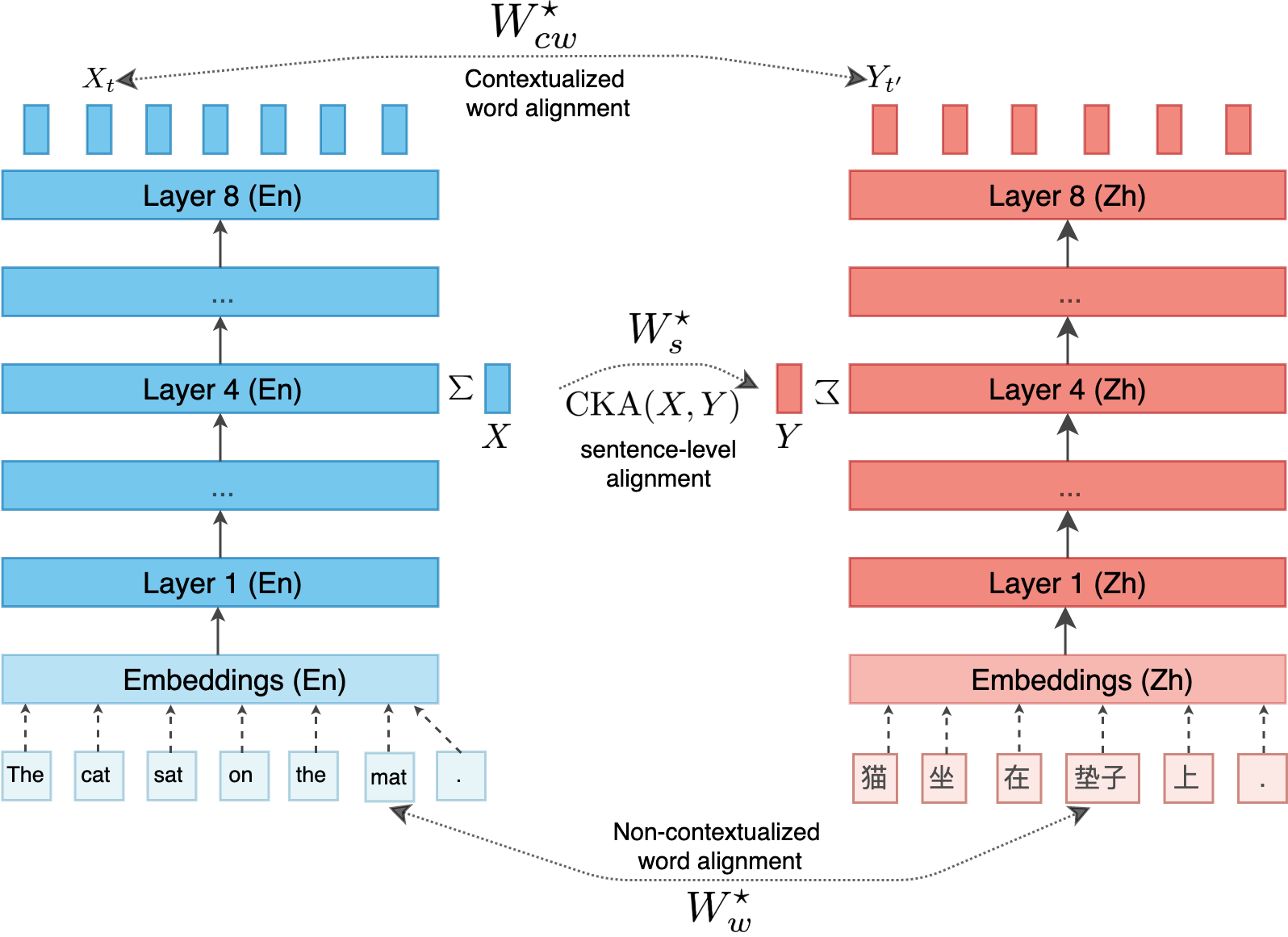}
        \caption{\small \textbf{Probing the layer similarity of monolingual BERT models.} We investigate the similarity of separate monolingual BERT models at different levels. We use an orthogonal mapping between the pooled representations of each model. We also quantify the similarity using the centered kernel alignment (CKA) similarity index.}
            \label{fig:params-2}
	\end{center}
    \end{minipage}
\end{figure*}

\section{Cross-lingual Pretraining}
We study a standard multilingual masked language modeling formulation and evaluate performance on several different cross-lingual transfer tasks, as described in this section. 

\subsection{Multilingual Masked Language Modeling}
Our multilingual masked language models follow the setup used by both mBERT and XLM. We use the implementation of \citet{lample2019cross}. Specifically, we consider continuous streams of 256 tokens and mask 15\% of the input tokens which we replace 80\% of the time by a mask token, 10\% of the time with the original word, and 10\% of the time with a random word. Note the random words could be foreign words. The model is trained to recover the masked tokens from its context~\cite{taylor1953cloze}. The subword vocabulary and model parameters are shared across languages. Note the model has a softmax prediction layer shared across languages. We use Wikipedia for training data, preprocessed by Moses \cite{koehn-etal-2007-moses} and Stanford word segmenter (for Chinese only) and BPE~\cite{sennrich-etal-2016-neural} to learn subword vocabulary. During training, we sample a batch of continuous streams of text from one language proportionally to the fraction of sentences in each training corpus, exponentiated to the power $0.7$. 

\paragraph{Pretraining details} Each model is a Transformer \cite{transformer17} with 8 layers, 12 heads and GELU activiation functions~\cite{hendrycks2016gaussian}. The output softmax layer is tied with input embeddings \cite{press-wolf-2017-using}. The embeddings dimension is 768, the hidden dimension of the feed-forward layer is 3072, and dropout is 0.1. We train our models with the Adam optimizer \cite{kingma2014adam} and the inverse square root learning rate scheduler of \newcite{transformer17} with $10^{-4}$ learning rate and 30k linear warmup steps. For each model, we train it with 8 NVIDIA V100 GPUs with 32GB of memory and mixed precision. It takes around 3 days to train one model. We use batch size 96 for each GPU and each epoch contains 200k batches. We stop training at epoch 200 and select the best model based on English dev perplexity for evaluation.

\subsection{Cross-lingual Evaluation}
\label{sec:eval}
We consider three NLP tasks to evaluate performance: natural language inference (NLI), named entity recognition (NER) and dependency parsing (Parsing). We adopt the \textbf{zero-shot cross-lingual transfer} setting, where we (1) fine-tune the pretrained model on English and (2) directly transfer the model to target languages. We select the model and tune hyperparameters with the English dev set. We report the result on average of best two set of hyperparameters.

\paragraph{Fine-tuning details} We fine-tune the model for 10 epochs for NER and Parsing and 200 epochs for NLI. We search the following hyperparameter for NER and Parsing: batch size $\{16, 32\}$; learning rate $\{\text{2e-5},\text{3e-5},\text{5e-5}\}$. For XNLI, we search: batch size $\{4, 8\}$; encoder learning rate $\{\text{1.25e-6},\text{2.5e-6},\text{5e-6}\}$; classifier learning rate $\{\text{5e-6},\text{2.5e-5},\text{1.25e-4}\}$. We use Adam with fixed learning rate for XNLI and warmup the learning rate for the first 10\% batch then decrease linearly to 0 for NER and Parsing. We save checkpoint after each epoch.

\paragraph{NLI} We use the cross-lingual natural language inference (XNLI) dataset \cite{conneau-etal-2018-xnli}. The task-specific layer is a linear mapping to a softmax classifier, which takes the representation of the first token as input. 

\paragraph{NER} We use WikiAnn \cite{pan-etal-2017-cross}, a silver NER dataset built automatically from Wikipedia, for English-Russian and English-French. For English-Chinese, we use CoNLL 2003 English \cite{tjong-kim-sang-de-meulder-2003-introduction} and a Chinese NER dataset~\cite{levow-2006-third}, with realigned Chinese NER labels based on the Stanford word segmenter. We model NER as BIO tagging. The task-specific layer is a linear mapping to a softmax classifier, which takes the representation of the first subword of each word as input. We report span-level F1. We adopt a simple post-processing heuristic to obtain a valid span, rewriting standalone \texttt{I-X} into \texttt{B-X} and \texttt{B-X I-Y I-Z} into \texttt{B-Z I-Z I-Z}, following the final entity type. We report the span-level F1.

\paragraph{Parsing} Finally, we use the Universal Dependencies (UD v2.3) \cite{ud-v2.3} for dependency parsing. We consider the following four treebanks: English-EWT, French-GSD, Russian-GSD, and Chinese-GSD. The task-specific layer is a graph-based parser \cite{dozat2016deep}, using representations of the first subword of each word as inputs. We measure performance with the labeled attachment score (LAS).

\section{Dissecting mBERT/XLM models}

\begin{figure*}[t]
\centering
\includegraphics[width=1.9\columnwidth]{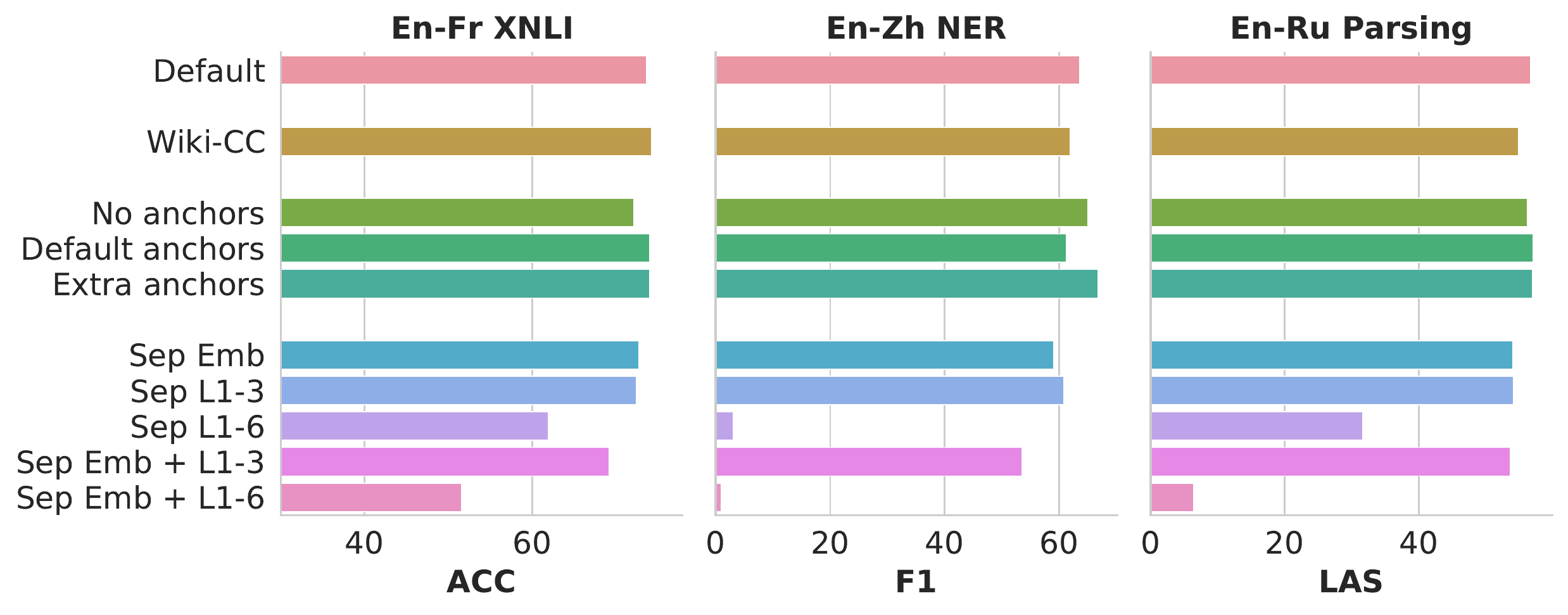}

\caption{Cross-lingual transfer of bilingual MLM on three tasks and language pairs under different settings.
Others tasks and languages pairs follows similar trend. See \cref{tab:all} for full results.}\label{fig:all}

\end{figure*}

We hypothesize that the following factors play important roles in what makes multilingual BERT multilingual: domain similarity, shared vocabulary (or anchor points), shared parameters, and language similarity. Without loss of generality, we focus on bilingual MLM. We consider three pairs of languages: English-French, English-Russian, and English-Chinese. 

\insertAllTable

\subsection{Domain Similarity}
\label{sec:domain-similarity}

Multilingual BERT and XLM are trained on the Wikipedia comparable corpora. Domain similarity has been shown to affect the quality of cross-lingual word embeddings~\cite{conneau2017word}, but this effect is not well established for masked language models. We consider domain difference by training on Wikipedia for English and a random subset of Common Crawl of the same size for the other languages (\textbf{Wiki-CC}). We also consider a model trained with Wikipedia only (\textbf{Default}) for comparison.

The first group in \cref{tab:all} shows domain mismatch has a relatively modest effect on performance. XNLI and parsing performance drop around 2 points while NER drops over 6 points for all languages on average. One possible reason is that the labeled WikiAnn data for NER consists of Wikipedia text; domain differences between source and target language during pretraining hurt performance more. Indeed for English and Chinese NER, where neither side comes from Wikipedia, performance only drops around 2 points.

\subsection{Anchor points}
\label{sec:anchor-points}

Anchor points are \textit{identical strings} that appear in both languages in the training corpus. Translingual words like \textit{DNA} or \textit{Paris} appear in the Wikipedia of many languages with the same meaning. In mBERT, anchor points are naturally preserved due to joint BPE and shared vocabulary across languages. Anchor point existence has been suggested as a key ingredient for effective cross-lingual transfer since they allow the shared encoder to have at least some direct tying of meaning across different languages~\cite{lample2019cross,pires-etal-2019-multilingual,wu-dredze-2019-beto}. However, this effect has not been carefully measured. 

We present a controlled study of the impact of anchor points on cross-lingual transfer performance by varying the amount of shared subword vocabulary across languages. 
Instead of using a single joint BPE with 80k merges, we use language-specific BPE with 40k merges for each language. We then build vocabulary by taking the union of the vocabulary of two languages and train a bilingual MLM (\textbf{Default anchors}). To remove anchor points, we add a language prefix to each word in the vocabulary before taking the union. Bilingual MLM (\textbf{No anchors}) trained with such data has no shared vocabulary across languages. However, it still has a single softmax prediction layer shared across languages and tied with input embeddings.

As \newcite{wu-dredze-2019-beto} suggest there may also be correlation between cross-lingual performance and anchor points,
we additionally increase anchor points by using a bilingual dictionary to create code switch data for training bilingual MLM (\textbf{Extra anchors}). For two languages, $\ell_1$ and $\ell_2$, with bilingual dictionary entries $d_{\ell_1, \ell_2}$, we add anchors to the training data as follows. 
For each training word $w_{\ell_1}$ in the bilingual dictionary, we either leave it as is (70\% of the time) or randomly replace it with one of the possible translations from the dictionary (30\% of the time).
We change at most 15\% of the words in a batch and sample word translations from  PanLex~\cite{kamholz-etal-2014-panlex} bilingual dictionaries, weighted according to their translation quality
\footnote{Although we only consider pairs of languages, this procedure naturally scales to multiple languages, which could produce larger gains in future work.}.

The second group of \cref{tab:all} shows cross-lingual transfer performance under the three anchor point conditions. Anchor points have a clear effect on performance and more anchor points help, especially in the less closely related language pairs (e.g. English-Chinese has a larger effect than English-French with over 3 points improvement on NER and XNLI). However, surprisingly, effective transfer is still possible with no anchor points. Comparing no anchors and default anchors, the performance of XNLI and parsing drops only around 1 point while NER even improve 1 points averaging over three languages.
Overall, these results show that we have previously overestimated the contribution of anchor points during multilingual pretraining.
Concurrently, \newcite{karthikeyan2020cross} similarly find anchor points play minor role in learning cross-lingual representation.

\subsection{Parameter sharing}
\label{sec:param-sharing}

Given that anchor points are not required for transfer, a natural next question is the extent to which we need to tie the parameters of the transformer layers. Sharing the parameters of the top layer is necessary to provide shared inputs to the task-specific layer. However, 
as seen in \cref{fig:params-1},
we can progressively separate the \textit{bottom} layers 1:3 and 1:6 of the Transformers and/or the embedding layers (including positional embeddings) (\textbf{Sep Emb}; \textbf{Sep L1-3}; \textbf{Sep L1-6}; \textbf{Sep Emb + L1-3}; \textbf{Sep Emb + L1-6}). Since the prediction layer is tied with the embeddings layer, separating the embeddings layer also introduces a language-specific softmax prediction layer for the cloze task. Additionally, we only sample random words within one language during the MLM pretraining. During fine-tuning on the English training set, we freeze the language-specific layers and only fine-tune the shared layers.

The third group in \cref{tab:all} shows cross-lingual transfer performance under different parameter sharing conditions with ``Sep'' denote which layers \textbf{is not} shared across languages. Sep Emb (effectively no anchor point) drops more than No anchors with 3 points on XNLI and around 1 point on NER and parsing, suggesting have a cross-language softmax layer also helps to learn cross-lingual representations. Performance degrades as fewer layers are shared for all pairs, and again the less closely related language pairs lose the most. Most notably, the cross-lingual transfer performance drops to random when separating embeddings and bottom 6 layers of the transformer. However, reasonably strong levels of transfer are still possible without tying the bottom three layers. These trends suggest that parameter sharing is the key ingredient that enables the learning of an effective cross-lingual representation space, and having language-specific capacity does not help learn a language-specific encoder for cross-lingual representation. Our hypothesis is that the representations that the models learn for different languages are similarly shaped and models can reduce their capacity budget by aligning representations for text that has similar meaning across languages.

\subsection{Language Similarity}

Finally, in contrast to many of the experiments above, language similarity seems to be quite important for effective transfer.  
Looking at \cref{tab:all} column by column in each task, we observe performance drops as language pairs become more distantly related. Using extra anchor points helps to close the gap. However, the more complex tasks seem to have larger performance gaps and having language-specific capacity does not seem to be the solution. Future work could consider scaling the model with more data and cross-lingual signal to close the performance gap.

\subsection{Conclusion}

Summarised by \cref{fig:all}, parameter sharing is the most important factor. More anchor points help but anchor points and shared softmax projection parameters are not necessary for effective cross-lingual transfer. Joint BPE and domain similarity contribute a little in learning cross-lingual representation.

\section{Similarity of BERT Models}

To better understand the robust transfer effects of the last section, we show that independently trained monolingual BERT models learn representations that are similar across languages, much like the widely observed similarities in word embedding spaces.
In this section, we show that independent monolingual BERT models produce highly similar representations when evaluated at the word level (\cref{sec:align-noncontextual-word}), contextual word-level (\cref{sec:align-contextual-word}),
and sentence level (\cref{sec:align-sentence}) . We also plot the cross-lingual similarity of neural network activation with center kernel alignment (\cref{sec:cka}) at each layer. We consider five languages: English, French, German, Russian, and Chinese.

\subsection{Aligning Monolingual BERTs}\label{sec:align}

To measure similarity, we learn an orthogonal mapping using the Procrustes~\cite{smith2017offline} approach:
$$
W^{\star} = \underset{W \in O_d(\mathbb{R})}{\text{argmin}} \Vert W X - Y \Vert_{F} = UV^T
$$
with $U\Sigma V^T = \text{SVD}(Y X^T)$, where $X$ and $Y$ are representation of two monolingual BERT models, sampled at different granularities as described below. We apply iterative normalization on $X$ and $Y$ before learning the mapping \cite{zhang-etal-2019-girls}.

\subsubsection{Word-level alignment}\label{sec:align-noncontextual-word}

\begin{figure*}[t]
\centering
\subfloat[][Non-contextual word embeddings alignment]{
\includegraphics[width=1\columnwidth]{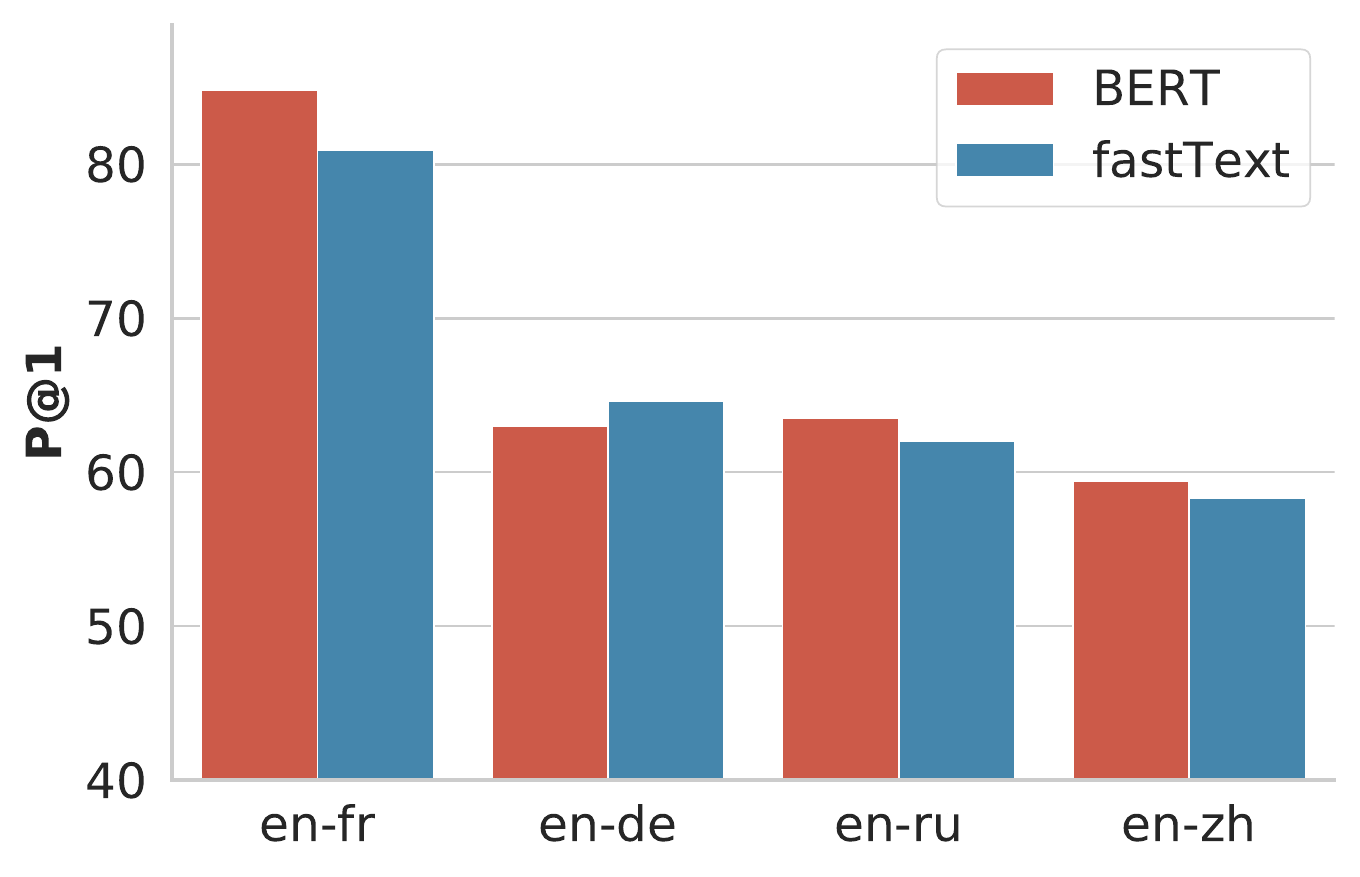}
\label{fig:align-noncontextual-word}}
\subfloat[][Contextual word embedding alignment]{
\includegraphics[width=1\columnwidth]{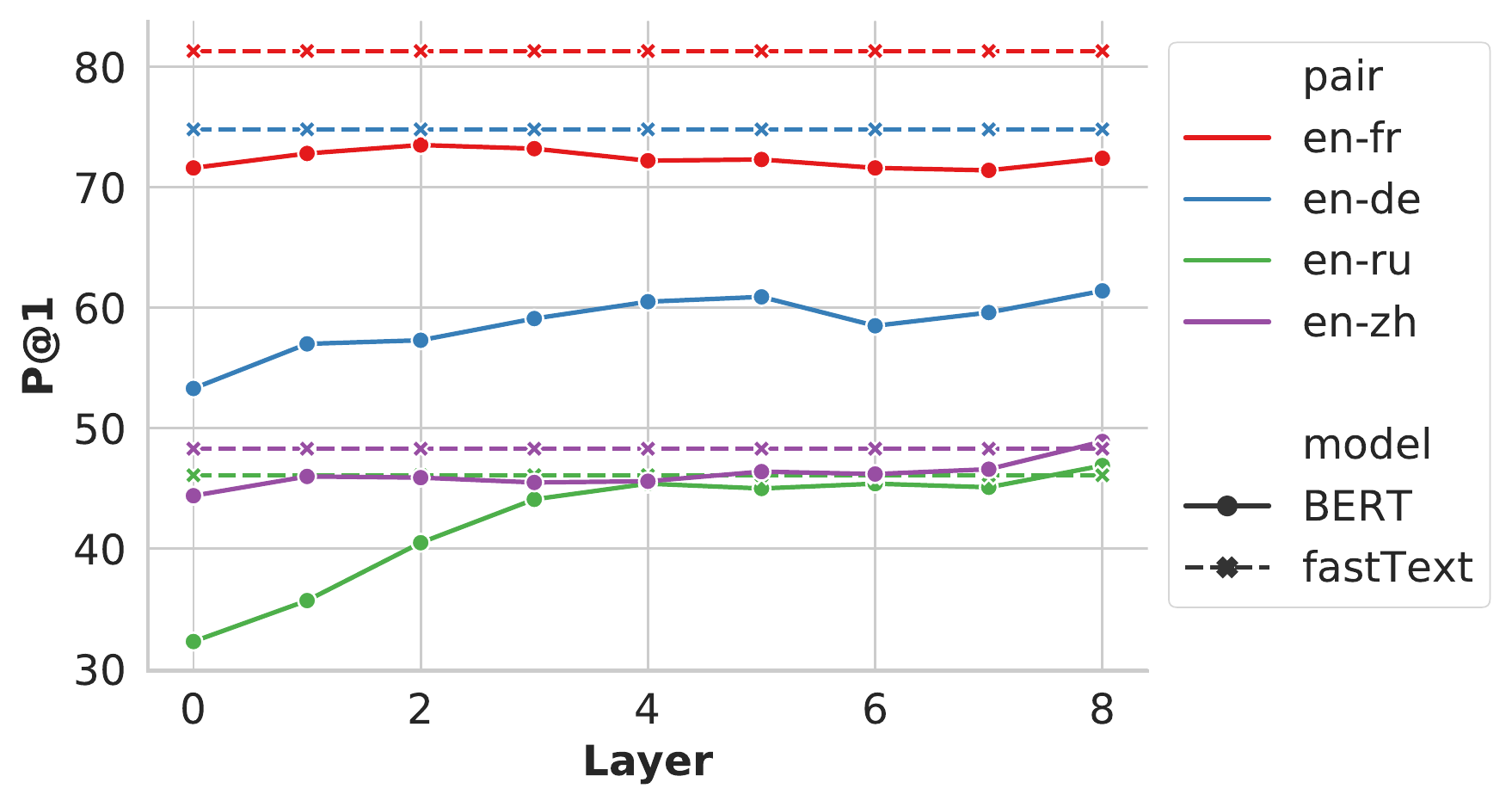}
\label{fig:align-contextual-word}}
\caption{Alignment of word-level representations from monolingual BERT models on subset of MUSE benchmark. \cref{fig:align-noncontextual-word} and \cref{fig:align-contextual-word} are not comparable due to different embedding vocabularies.}\label{fig:align-word}

\end{figure*}

\begin{figure*}[t]
\centering
\includegraphics[width=2\columnwidth]{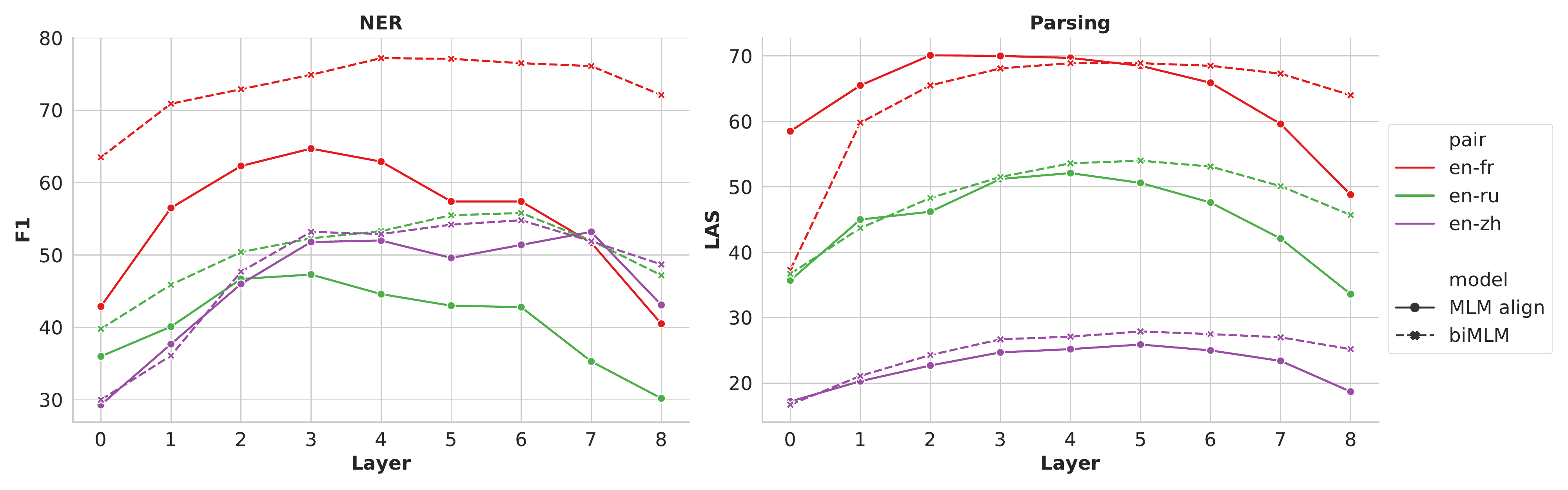}

\caption{Contextual representation alignment of different layers for zero-shot cross-lingual transfer.}\label{fig:align-contextual}
\vspace{-0.4cm}
\end{figure*}

In this section, we align both the non-contextual word representations from the embedding layers, and the contextual word representations from the hidden states of the Transformer at each layer.

For non-contextualized word embeddings, we define $X$ and $Y$ as the word embedding layers of monolingual BERT, which contain a single embedding per word (type). Note that in this case we only keep words containing only one subword. For contextualized word representations, we first encode 500k sentences in each language. At each layer, and for each word, we collect all contextualized representations of a word in the 500k sentences and average them to get a single embedding. Since BERT operates at the subword level, for one word we consider the average of all its subword embeddings. Eventually, we get one word embedding per layer.  We use the MUSE benchmark \cite{conneau2017word}, a bilingual dictionary induction dataset for alignment supervision and evaluate the alignment on word translation retrieval.
As a baseline, we use the first 200k embeddings of fastText \cite{bojanowski-etal-2017-enriching} and learn the mapping using the same procedure as \cref{sec:align}. Note we use a subset of 200k vocabulary of fastText, the same as BERT, to get a comparable number. We retrieve word translation using CSLS \cite{conneau2017word} with K=10.

In \cref{fig:align-word}, we report the alignment results under these two settings. \cref{fig:align-noncontextual-word} shows that the subword embeddings matrix of BERT, where each subword is a standalone word, can easily be aligned with an orthogonal mapping and obtain slightly better performance than the same subset of fastText. \cref{fig:align-contextual-word} shows embeddings matrix with the average of all contextual embeddings of each word can also be aligned to obtain a decent quality bilingual dictionary, although underperforming fastText. We notice that using contextual representations from higher layers obtain better results compared to lower layers.

\subsubsection{Contextual word-level alignment}\label{sec:align-contextual-word}

In addition to aligning word representations, we also align representations of two monolingual BERT models in contextual setting, and evaluate performance on cross-lingual transfer for NER and parsing. We take the Transformer layers of each monolingual model up to layer $i$, and learn a mapping $W$ from layer $i$ of the target model to layer $i$ of the source model. To create that mapping, we use the same Procrustes approach but use a dictionary of parallel contextual words, obtained by running the fastAlign~\cite{dyer-etal-2013-simple} model on the 10k XNLI parallel sentences. %

For each downstream task, we learn task-specific layers on top of $i$-th English layer: four Transformer layers and a task-specific layer. We learn these on the training set, but keep the first $i$ pretrained layers freezed. After training these task-specific parameters, we encode (say) a Chinese sentence with the first $i$ layers of the target Chinese BERT model, project the contextualized representations back to the English space using the $W$ we learned, and then use the task-specific layers for NER and parsing.

In \cref{fig:align-contextual}, we vary $i$ from the embedding layer (layer 0) to the last layer (layer 8) and present the results of our approach on parsing and NER. We also report results using the first $i$ layers of a bilingual MLM (biMLM).
\footnote{In \cref{app:bimlm-align}, we also present the same alignment step with biMLM but only observed improvement in parsing.}
We show that aligning monolingual models (MLM align) obtain relatively good performance even though they perform worse than bilingual MLM, except for parsing on English-French.
The results of monolingual alignment generally shows that we can align contextual representations of monolingual BERT models with a simple linear mapping and use this approach for cross-lingual transfer. We also observe that the model obtains the highest transfer performance with the middle layer representation alignment, and not the last layers. The performance gap between monolingual MLM alignment and bilingual MLM is higher in NER compared to parsing, suggesting the syntactic information needed for parsing might be easier to align with a simple mapping while entity information requires more explicit entity alignment.

\subsubsection{Sentence-level alignment}\label{sec:align-sentence}

In this case, $X$ and $Y$ are obtained by average pooling subword representation (excluding special token) of sentences \textit{at each layer} of monolingual BERT. We use multi-way parallel sentences from XNLI for alignment supervision and Tatoeba \cite{schwenk2019wikimatrix} for evaluation.

\begin{figure}[h!]
\begin{center}
\includegraphics[width=\columnwidth]{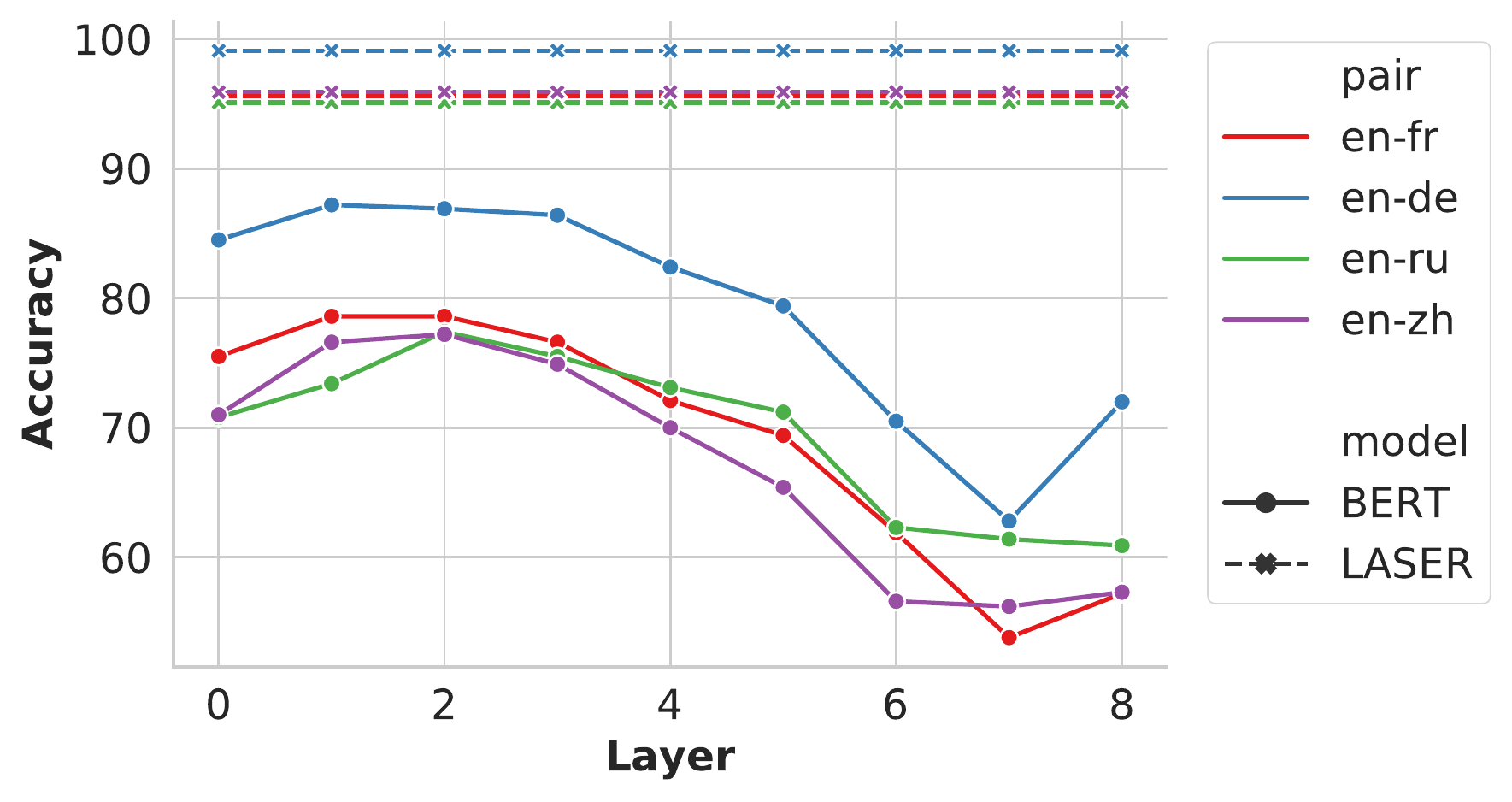}
\caption{Parallel sentence retrieval accuracy after Procrustes alignment of monolingual BERT models.
\label{fig:tatoeba}}
\end{center}
\end{figure}

\cref{fig:tatoeba} shows the sentence similarity search results with nearest neighbor search and cosine similarity, evaluated by precision at 1, with four language pairs. Here the best result is obtained at lower layers. The performance is surprisingly good given we only use 10k parallel sentences to learn the alignment without fine-tuning at all. As a reference, the state-of-the-art performance is over 95\%, obtained by LASER \cite{artetxe2019massively} trained with millions of parallel sentences.

\begin{figure*}[th]
\centering
\includegraphics[width=1.8\columnwidth]{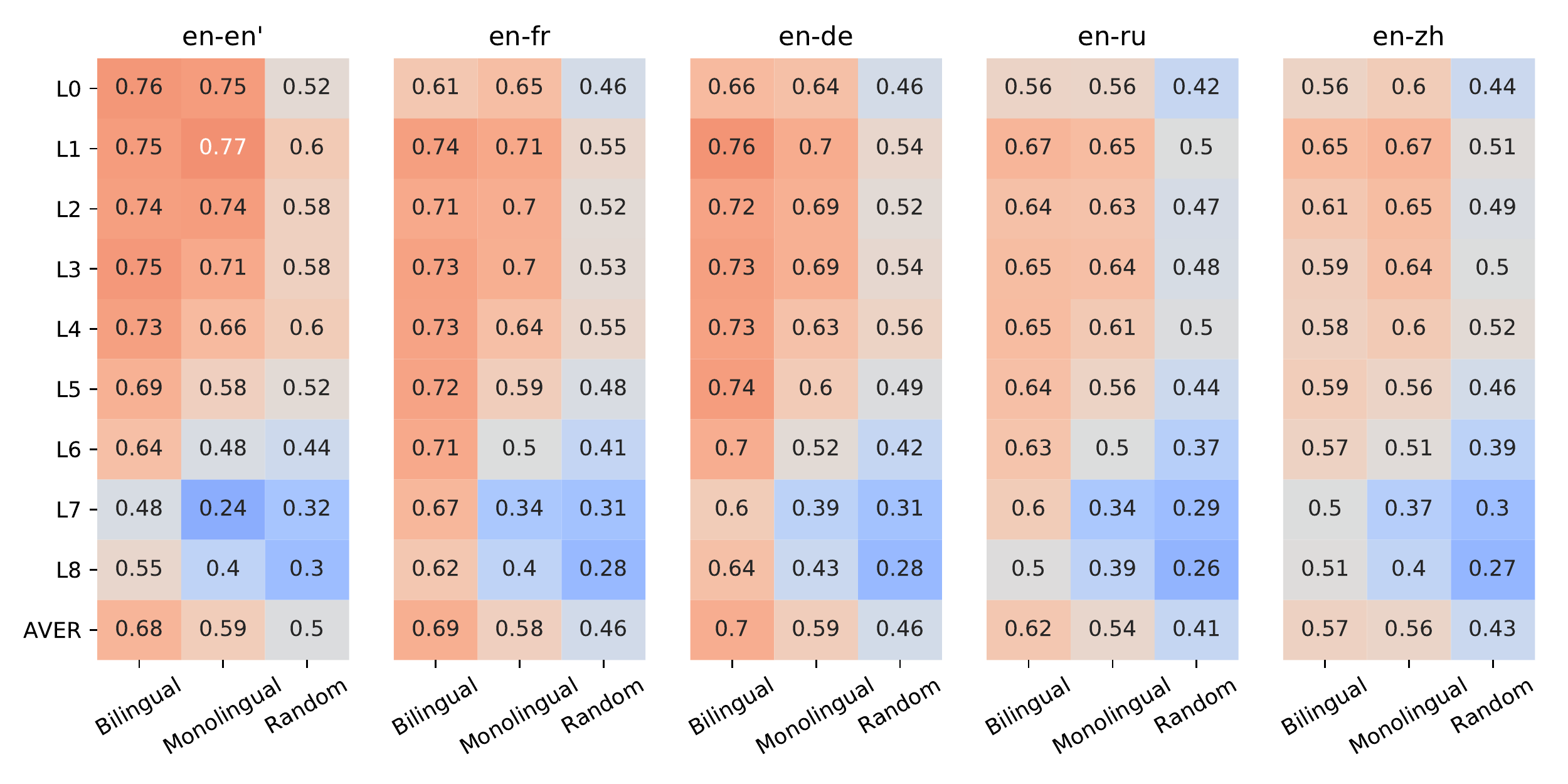}
\caption{CKA similarity of mean-pooled multi-way parallel sentence representation at each layers. Note en$^\prime$ corresponds to paraphrases of en obtained from back-translation (en-fr-en$^\prime$). Random encoder is only used by non-Engligh sentences. L0 is the embeddings layers while L1 to L8 are the corresponding transformer layers. The average row is the average of 9 (L0-L8) similarity measurements.}\label{fig:cka}
\vspace{-0.3cm}
\end{figure*}

\subsubsection{Conclusion} These findings demonstrate that both word-level, contextual word-level, and sentence-level BERT representations can be aligned with a simple orthogonal mapping. Similar to the alignment of word embeddings~\cite{mikolov2013exploiting}, this shows that BERT models are similar across languages. This result gives more intuition on why mere parameter sharing is sufficient for multilingual representations to emerge in multilingual masked language models.

\subsection{Neural network similarity}\label{sec:cka}

Based on the work of \newcite{kornblith2019similarity}, we examine the centered kernel alignment (CKA), a neural network similarity index that improves upon canonical correlation analysis (CCA), and use it to measure the similarity across both monolingual and bilingual masked language models.
The linear CKA is both invariant to orthogonal transformation and isotropic scaling, but are not invertible to any linear transform.
The linear CKA similarity measure is defined as follows:
$$
\text{CKA}(X,Y) = \frac{\Vert Y^TX \Vert^2_{\text{F}}}{(\Vert X^TX\Vert_{\text{F}} \Vert Y^TY\Vert_{\text{F}})},
$$
where $X$ and $Y$ correspond respectively to the matrix of the $d$-dimensional mean-pooled (excluding special token) subword representations at layer $l$ of the $n$ parallel source and target sentences.

In \cref{fig:cka}, we show the CKA similarity of monolingual models, compared with bilingual models and random encoders, of multi-way parallel sentences \cite{conneau-etal-2018-xnli} for five languages pair: English to English$^\prime$ (obtained by back-translation from French), French, German, Russian, and Chinese. The monolingual en$^\prime$ is trained on the same data as en but with different random seed and the bilingual en-en$^\prime$ is trained on English data but with separate embeddings matrix as in \cref{sec:param-sharing}. The rest of the bilingual MLM is trained with the Default setting. We only use random encoder for non-English sentences.

\cref{fig:cka} shows bilingual models have slightly higher similarity compared to monolingual models with random encoders serving as a lower bound. Despite the slightly lower similarity between monolingual models, it still explains the alignment performance in \cref{sec:align}. Because the measurement is also invariant to orthogonal mapping, the CKA similarity is highly correlated with the sentence-level alignment performance in \cref{fig:tatoeba} with over 0.9 Pearson correlation for all four languages pairs. For monolingual and bilingual models, the first few layers have the highest similarity, which explains why \newcite{wu-dredze-2019-beto} finds freezing bottom layers of mBERT helps cross-lingual transfer. The similarity gap between monolingual model and bilingual model decrease as the languages pair become more distant. In other words, when languages are similar, using the same model increase representation similarity. On the other hand, when languages are dissimilar, using the same model does not help representation similarity much. Future work could consider how to best train multilingual models covering distantly related languages.

\section{Discussion}
In this paper, we show that multilingual representations can emerge from unsupervised multilingual masked language models with only parameter sharing of some Transformer layers. Even without any anchor points, the model can still learn to map representations coming from different languages in a single shared embedding space. We also show that isomorphic embedding spaces emerge from monolingual masked language models in different languages, similar to word2vec embedding spaces~\cite{mikolov2013exploiting}. By using a linear mapping, we are able to align the embedding layers and the contextual representations of Transformers trained in different languages. We also use the CKA neural network similarity index to probe the similarity between BERT Models and show that the early layers of the Transformers are more similar across languages than the last layers. All of these effects were stronger for more closely related languages, suggesting there is room for significant improvements on more distant language pairs. 

\bibliography{anthology,acl2020}
\bibliographystyle{acl_natbib}

\clearpage
\appendix

\section{Contextual word-level alignment of bilingual MLM representation}\label{app:bimlm-align}

\begin{figure}[h]
\centering
\includegraphics[width=\columnwidth]{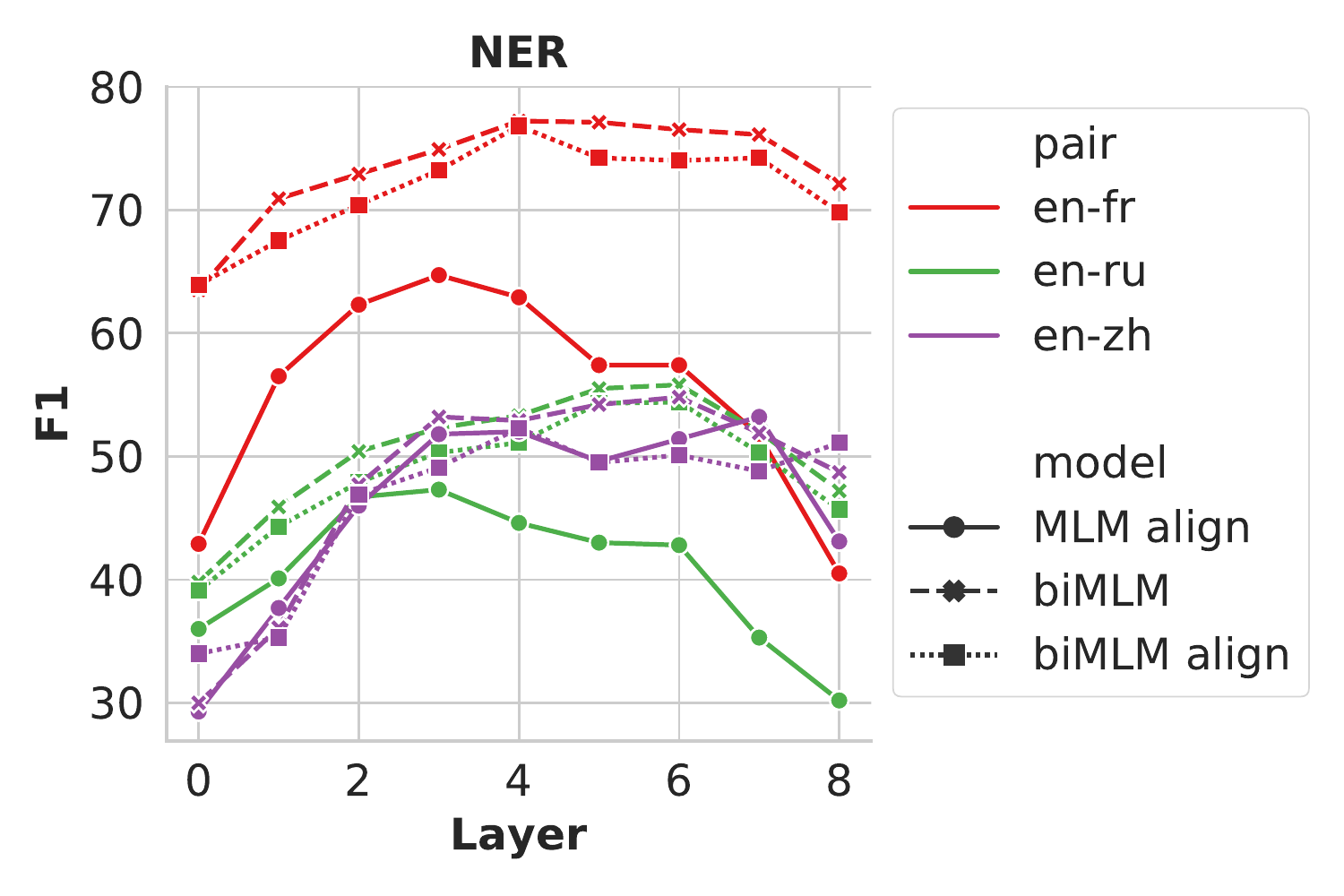}

\includegraphics[width=\columnwidth]{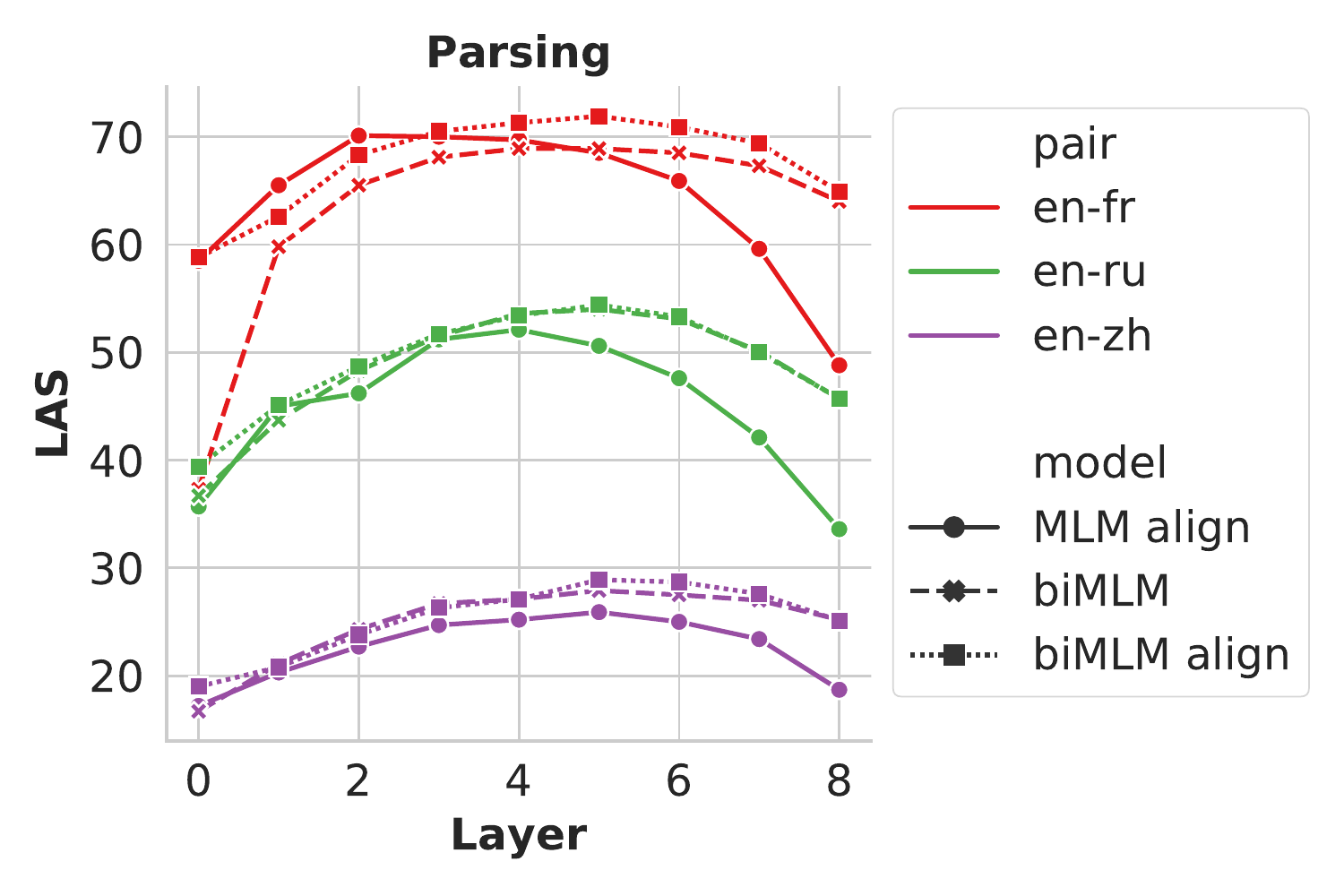}

\caption{Contextual representation alignment of different layers for zero-shot cross-lingual transfer.}
\end{figure}

\end{document}